%% file: main.tex
\pdfoutput=1

\documentclass[11pt]{article}

\usepackage{acl}
\usepackage{times}
\usepackage{latexsym}
\usepackage{graphicx}

\usepackage[T1]{fontenc}
\usepackage[utf8]{inputenc}
\usepackage{microtype}
\usepackage{inconsolata}

\title{SparkRA: A Retrieval-Augmented Knowledge Service System Based on Spark Large Language Model}

\author{Dayong Wu\textsuperscript{1},
Jiaqi Li\textsuperscript{1,2},
Baoxin Wang\textsuperscript{1,3}, 
Honghong Zhao\textsuperscript{1}, 
Siyuan Xue\textsuperscript{1}, 
Yanjie Yang\textsuperscript{1}, 
Zhijun Chang\textsuperscript{4},\\
\textbf{Rui Zhang}\textsuperscript{1}, 
\textbf{Li Qian}\textsuperscript{4}, 
\textbf{Bo Wang}\textsuperscript{1}, 
\textbf{Shijin Wang}\textsuperscript{1}, 
\textbf{Zhixiong Zhang}\textsuperscript{4}, 
\textbf{Guoping Hu}\textsuperscript{1}\\
1. State Key Laboratory of Cognitive Intelligence, iFLYTEK Research, China \\
2. University of Science and Technology of China, China. \\
3. Harbin Institute of Technology, China.\\
4. National Science Library, Chinese Academy of Sciences, China.
}

\begin{document}
\maketitle

\input{abstract}
\input{introduction}
\input{methodology}

\input{sparkra}

\input{experiments}

\input{related_work}

\input{conclusion}

\bibliography{main}

\end{document}

%% file: abstract.tex
\begin{abstract}

Large language models (LLMs) have shown remarkable achievements across various language tasks.
To enhance the performance of LLMs in scientific literature services, we developed the scientific literature LLM (SciLit-LLM) through pre-training and supervised fine-tuning on scientific literature, building upon the iFLYTEK Spark LLM.
Furthermore, we present a knowledge service system Spark Research Assistant (SparkRA) based on our SciLit-LLM. 
SparkRA is accessible online\footnote{https://paper.iflytek.com/} and provides three primary functions: literature investigation, paper reading, and academic writing.
As of July 30, 2024, SparkRA has garnered over 50,000 registered users, with a total usage count exceeding 1.3 million.

\end{abstract}

%% file: introduction.tex
\section{Introduction}

Large language models (LLMs) have achieved significant success in natural language processing, including text generation and language understanding \cite{Brown2020gpt-3,Chowdhery2023}. 
Owing to their strong capabilities, LLMs have shown immense potential across many downstream fields, such as education, medicine, and finance \cite{Kasneci2023,Thirunavukarasu2023, Clusmann2023,Shah2023}.


As the performance of LLMs in scientific literature does not fully meet the needs of scholars, we developed a Scientific Literature LLM (SciLit-LLM).
We began by collecting a large dataset of scientific literature, including academic papers and patents, and performed data cleaning to ensure high-quality academic text. We then continued pre-training the open-source iFLYTEK Spark LLM (13B)\footnote{https://gitee.com/iflytekopensource/iFlytekSpark-13B} using an autoregressive training task, followed by supervised fine-tuning, to create our SciLit-LLM.

\begin{figure}
    \centering
    \includegraphics[width=0.43\textwidth]{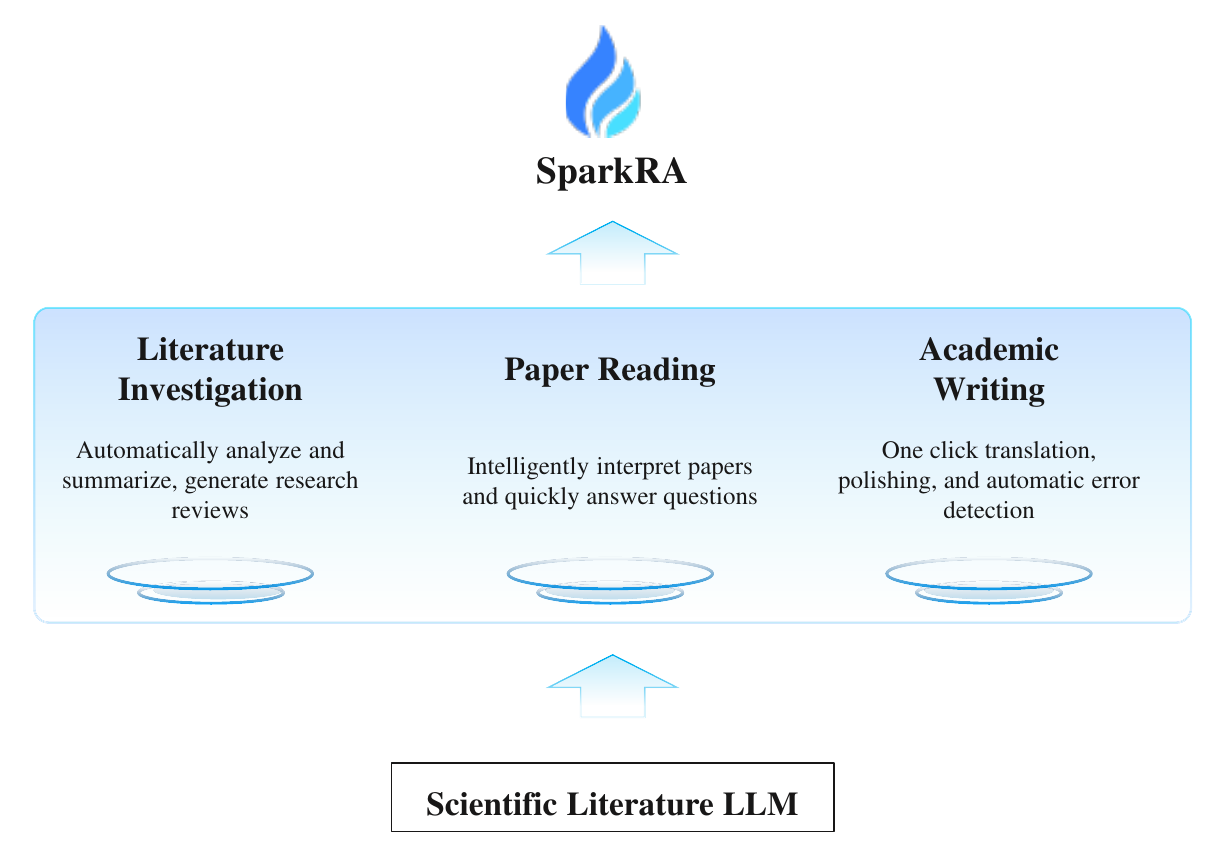}
    \caption{The process of building SparkRA system.}
    \label{fig:framework}
\end{figure}

Traditional knowledge service systems generally provide limited functionalities, such as the retrieval of scholarly articles and assistive reading services. In this paper, we introduce the Spark Research Assistant (SparkRA), a knowledge service system based on our scientific literature LLM. SparkRA offers a comprehensive, one-stop solution for scientific literature services. Figure 1 depicts the process of constructing the SparkRA system. The features of SparkRA are as follows:
\begin{itemize}
    \item Literature investigation: this sub-system can automatically analyze and summarize research areas, and generate research reviews.
    \item Paper reading: this sub-system can intelligently interpret papers and quickly answer questions. 
    \item Academic writing: this sub-system can provide the functions for writing academic papers including one-click translation, polishing, and automatic error detection.
\end{itemize}


Experimental evaluation demonstrates that SparkRA outperforms existing models, including GPT-3.5 and Llama3-8B, across all tasks, establishing its efficacy in enhancing the productivity and accuracy of academic research activities.

%% file: methodology.tex
\section{Scientific Literature LLM}

\subsection{Base model}

To build the LLM for scientific literature services, we selected the Spark LLM 
as the foundation model for building our scientific literature LLM (ScLlit-LLM). The Spark LLM, developed by iFLYTEK Research, demonstrates impressive performance in processing both English and Chinese languages. iFlytekSpark-13B has consistently ranked among the top in numerous well-known public benchmarks, demonstrating its superiority. Its performance is notably superior to other open-source models of equivalent size.

\subsection{Continual pre-training}

While the Spark LLM exhibits strong capabilities in language comprehension and text generation, it may struggle to directly provide accurate responses to scholarly inquiries without targeted training in the scientific domain. Consequently, we have designed a Scientific literature LLM that is specifically oriented towards parsing and understanding scientific literature.

Inspired by the existing research \cite{Beltagy2019scibert, Hong2022scholarbert}, we have further pre-trained the spark model on an extensive corpus of academic texts to enhance the model's performance in processing and generating scientific literature

\paragraph{Data preparation.} To enhance the foundational large language model (LLM), it is imperative to amass a vast corpus of high-quality data, which includes kinds of scholarly literature like papers and patents. We collected a vast number of academic papers from various publicly accessible websites, such as arXiv\footnote{https://arxiv.org/}.

Given that academic documents are predominantly archived in PDF format, it is crucial to convert these PDFs into text while meticulously eliminating any extraneous elements. For this purpose, we employed a sophisticated PDF parsing tool developed by iFLYTEK. In the process of advancing our scientific literature LLM, we have incorporated a dataset comprising over 10M academic papers.

To prevent LLM from losing its general capabilities, we also incorporated a significant amount of general corpora. This strategy ensures that after continual pre-training, the scientific literature LLM performs better in the field of science while maintaining the general capabilities.

\paragraph{Pre-training.} Similar to the traditional LLM  pre-training process, the scientific literature LLM employs the same next-word prediction task for its continual pre-training on a corpus of scientific literature comprising billions of tokens. 

Upon evaluation, the scientific literature LLM, continual pre-training, exhibits improved performance on general scholarly inquiries. Moreover, for specialized academic queries without provided context, the scientific literature LLM demonstrates a higher rejection tendency, effectively reducing instances of hallucination.

\subsection{Supervised fine-tuning}

Supervised fine-tuning (SFT) is a technique used to enhance large language models (LLMs) by further training a pre-trained model to improve its accuracy and relevance for specific tasks or domains. The efficacy of SFT in refining LLMs is well-documented \cite{Wei2022FLAN,ouyang2022instructgpt}. This process involves utilizing a carefully curated dataset with labeled examples that illustrate the desired output. During SFT, the model learns from these examples to comprehend the intricacies of the task more thoroughly. Consequently, SFT enables the model to retain its broad knowledge base while acquiring specialization in targeted areas, resulting in enhanced user experiences and more precise information delivery.


\paragraph{Data preparation.} In the construction of our datasets for supervised fine-tuning, each instance within datasets is composed of three elements: an instruction, an input, and an output. We utilize a dual approach in formulating instructions, leveraging both Self-instruct \cite{Wang2023selfinstruct} and human writing. 

\begin{figure*}
    \centering
    \includegraphics[width=0.8\textwidth]{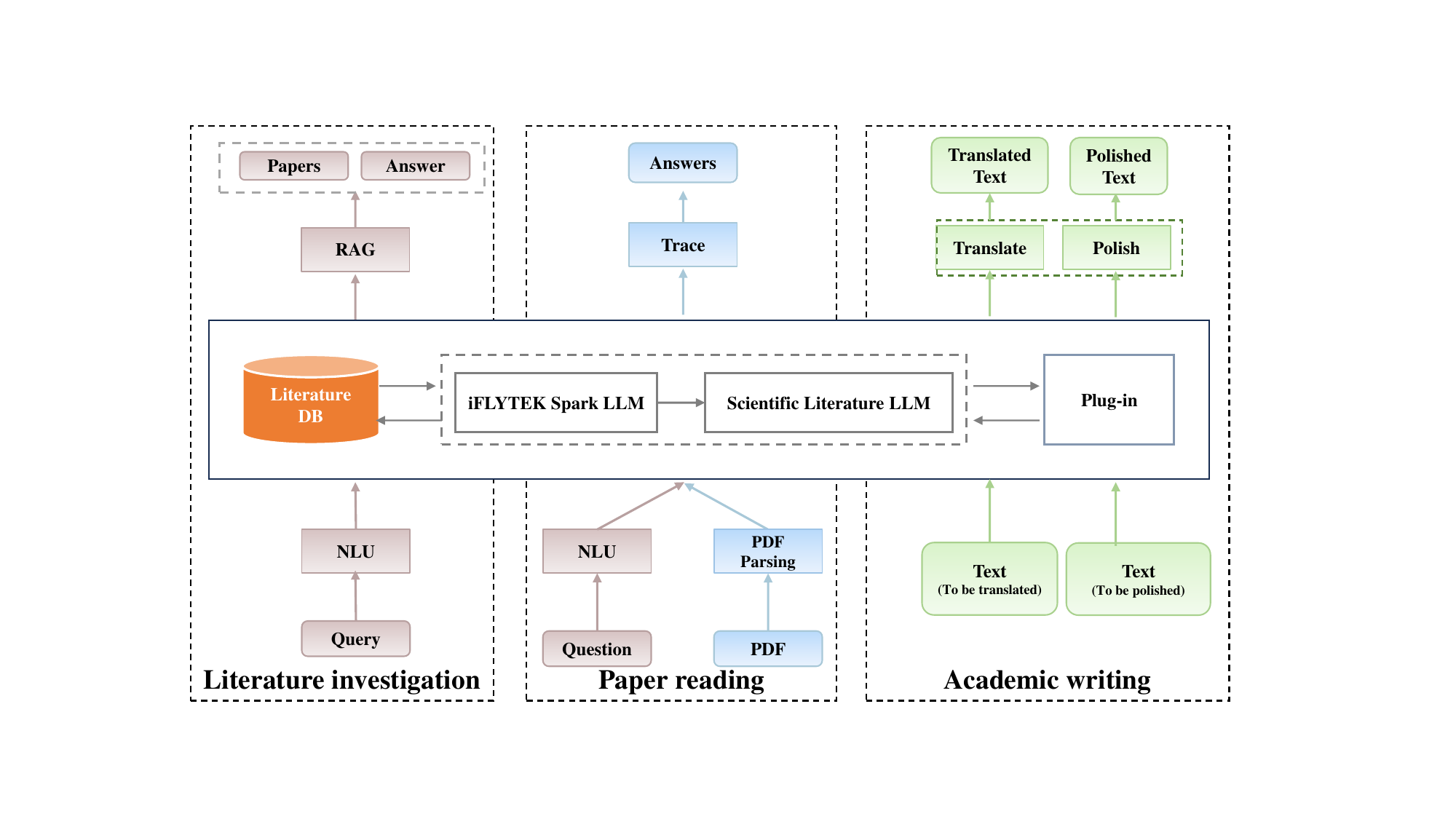}
    \caption{The system architecture of SparkRA integrates iFLYTEK Spark LLM and Scientific Literature LLM to facilitate literature investigation, paper reading, and academic writing.}
    \label{fig:framework}
\end{figure*}

To exemplify, consider the instruction: ``Please translate the input English sentence into Chinese''; here, the input component would be an English sentence. For the generation of outputs corresponding to given instructions and inputs, we employ meticulously devised manual methods to craft expert responses.

\paragraph{Training.} Upon completing the construction of SFT datasets, we commenced the Supervised Fine-Tuning (SFT) of scientific literature LLM. The instances within the dataset serve as labeled data for the SFT of the model. Since each instance is meticulously crafted by experts, they are of higher quality compared to the generic data used during the pre-training phase. Moreover, these labeled data enhance the LLM's ability to answer questions. The scientific literature LLM that has undergone SFT with domain-specific data can learn from experts' responses to research-related inquiries and generalize this knowledge to a broader array of questions.

%% file: sparkra.tex
\section{SparkRA}

Based on our SciLit-LLM, we developed a literature services system SparkRA. This platform is comprised of three functions: literature investigation, paper reading, and academic writing. Notably, SparkRA is equipped to process inputs in both Chinese and English, thereby catering to a diverse linguistic user base. The architecture of SparkRA is shown in Figure 2 and the demonstration video has been published on YouTube\footnote{https://youtu.be/bdUMTr3pMfY}.

\subsection{Literature investigation}

This function is designed to facilitate the exploration of academic literature and is comprised of three integral components: an investigation copilot, a research topic search engine, and a review generation module. The architecture and screenshot of the literature investigation function are respectively shown in Figure 3 and Figure 4.

\begin{figure}
    \centering
    \includegraphics[width=0.23\textwidth]{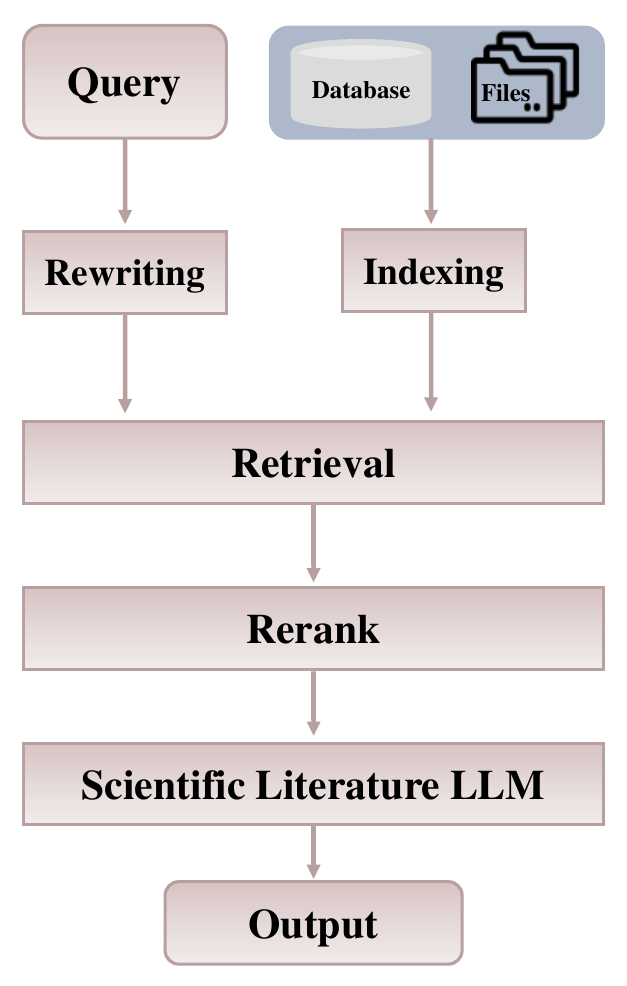}
    \caption{The architecture of RAG-based literature investigation.}
    \label{fig:framework}
\end{figure}

\paragraph{Investigation copilot.} This copilot assists users in deepening their understanding of specific research domains and various scholars through interactive natural language dialogue.

(1) Area-based survey. Users can easily obtain the summarization and papers of a specific research area. For example, the user can send the query ``What are the recent papers of fake news section in 2023''. SparkRA will show the papers and give a summary.

(2) Scholar-based survey. This function can output the papers of the input scholar and divide the papers into different research areas. For example, the user can send the query ``What research has Chris Manning from Stanford University conducted''.

\begin{figure*}
    \centering
    \includegraphics[width=0.9\textwidth]{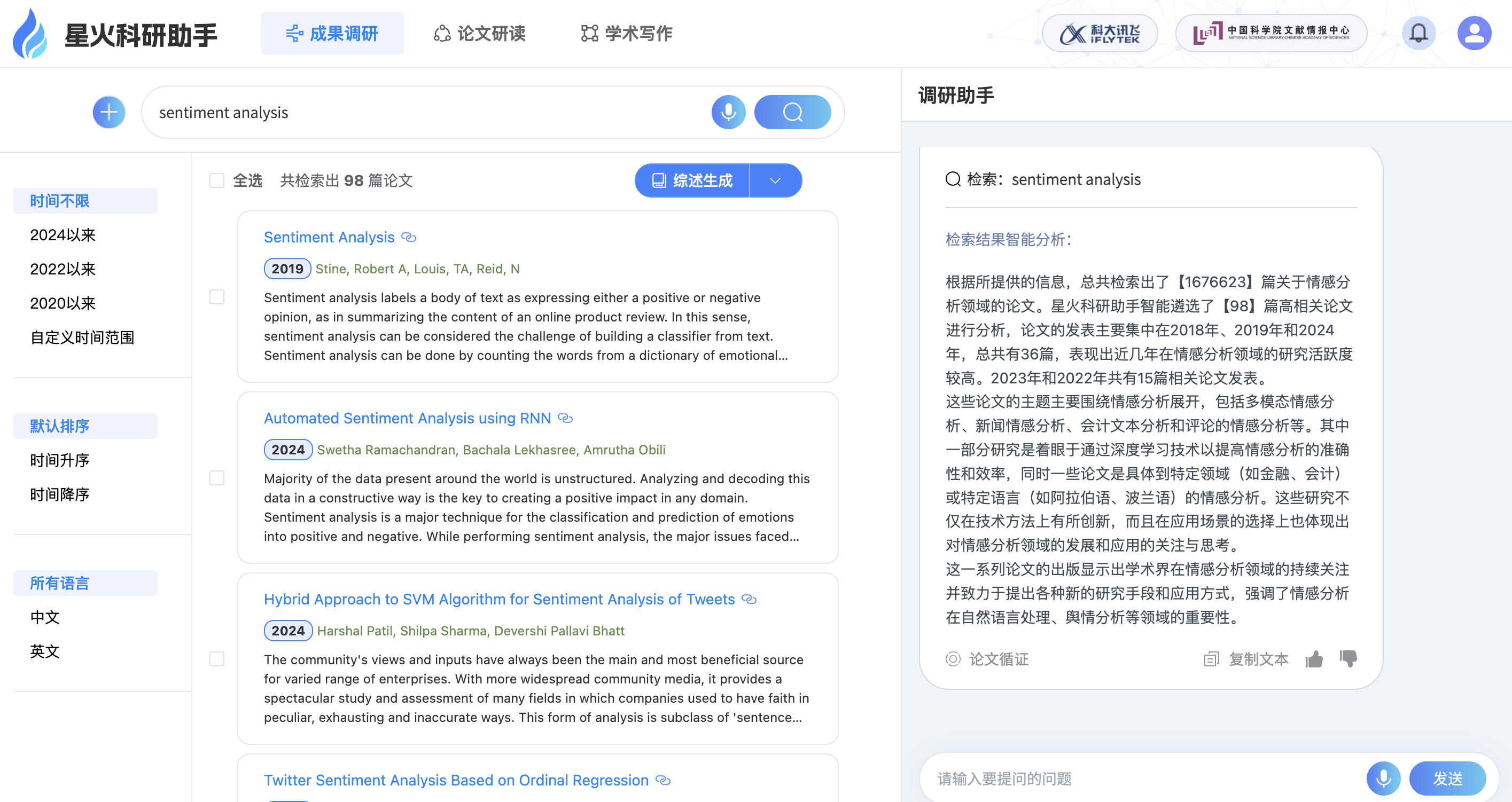}
    \caption{Literature investigation page.}
    \label{fig:framework}
\end{figure*}

\paragraph{Topic search engine.} The search interface accommodates queries pertaining to research topics in both Chinese and English. Upon receiving a specified topic, SparkRA retrieves relevant papers from an extensive academic library and provides concise summaries of their content.

(1) Query rewriting. There is a diversity of user retrieval query formats and the occasional inclusion of noise, such as ``In the library, what LLM technologies can assist users in improving the efficiency of finding books?''. Upon receiving a user's query, scientific literature LLM is used to revise the query into a format more suited for retrieval, like ``Applications of large models in library search domain''. This strategy can significantly enhance the system's ability to locate the desired literature.

(2) Precise Retrieval. Upon completion of the rewriting process, the revised query is subjected to information extraction through natural language understanding technologies, such as Named Entity Recognition (NER). The extracted information encompasses scholars, institutions, dates, domains, and keywords, among others. Based on the extracted content, the corresponding search plugin interfaces are invoked to obtain precise search results.

(3) Literature-based summary. Building on the retrieval outcomes, the scientific literature LLM synthesizes findings, encompassing the distribution of publication years, trends in literature popularity, recent focal topics, and potential future directions of development.

\paragraph{Review generation.} This function enables the generation of a report based on a selection of papers, with a maximum limit of 30 papers.
The generated report facilitates an expedited comprehension of a substantial volume of literature within a specific domain or authored by an individual.

In this function, we leveraged the clustering capabilities and inductive summarization prowess of LLM. Through the clustering of dozens of literature papers, the model structured the introduction, body, and conclusion of a comprehensive review, including the formulation of pertinent headings. Subsequently, the model demonstrated its robust capacity for inductive reasoning and summarization. It also featured the capability to annotate the analytical text with hyperlinks, serving as citations that facilitate reference validation at the end of the review and enable user verification.


\subsection{Paper reading}

This function can assist scholars and students in reading academic papers. With the rapid development of artificial intelligence technology, a large number of cutting-edge papers emerge every day. It is necessary to develop an intelligent system to help people understand papers. 

For paper reading, LLMs with longer context windows are required because the full article of paper is usually long. However, training an LLM with long context windows from scratch requires significantly larger investments. To facilitate this, we employ a retrieval-augmented approach to enhance the effectiveness of the large model's answers. We initiate text splitting as a primary step and engage in chapter recognition to preserve the semantic integrity of segments.
For the cross-language retrieval embedding model, firstly, we generate questions from paper segments using an LLM and construct a large set of (question, positive sample, negative samples) pairs for training. Subsequently, we use XLM-RoBERTa \cite{conneau2020xml-r} as the language encoder and fine-tune the model via contrastive learning. The input question and retrieved segments are finally fed into the SciLit-LLM to generate answers.

\textbf{Reading Copilot} enhances paper comprehension through natural language interactions. Questions fall into two categories: those within the paper, which SciLit-LLM answers using the input paper alone, and those outside the paper, which require a search engine plugin to retrieve relevant information. For the latter, answers are generated through retrieval-augmented generation using SciLit-LLM.


\textbf{Multi-Document Comparison} allows for the comparison of two to five papers. For each selected paper, SparkRA provides the abstract and contributions separately. It also generates a comparative analysis table that highlights the proposed approaches and advantages of each paper. SparkRA can identify and output both the similarities and differences among the selected papers.


\subsection{Academic writing}

This function is directly powered by SciLit-LLM and includes polishing and translation.

\paragraph{Paper polishing.} This function is used to assist the scholar and students in polishing the academic paper draft.  We construct a large corpus of texts requiring polishing based on a multitude of well-written academic papers, utilizing few-shot learning and chain-of-thought (COT) prompting methodologies, followed by supervised learning for instruction fine-tuning.

\paragraph{Academic translation.} 
In order to accurately translate domain-specific terminology, we have implemented a dynamic perception prompts approach to guide the model in completing translation tasks. Based on the user's input prompts, we obtain prompts with professional terminology translations from a terminology translation lexicon in the knowledge base, which are then fed into the large language model.

%% file: experiments.tex
\section{Experiments}

\subsection{Experiment setting}

To validate the results of SparkRA, we adopt the following LLMs as the baseline models:
\begin{itemize}
    \item Llama: a large-scale language model developed and open-sourced by Meta, was compared to SciLit-LLM using three versions: Llama2-7B, Llama2-13B, and Llama3-8B.
    \item ChatGPT (GPT-3.5): it is a large-scale language model in the field of artificial intelligence developed by OpenAI. 
    \item GPT-4: GPT-4 Turbo serves as our baseline model, consistently outperforming in a range of NLP tasks.
\end{itemize}


We evaluate the performance of models using the mean opinion score (MOS) on a scale of 1 (poorest) to 5 (optimal), with evaluations conducted by more than five individuals per task. For the machine translation task, we also use the BLEU metric \cite{bleu2002} for model evaluation. We gathered 100 academic parallel paragraphs from public Chinese journals with Chinese and English abstracts to serve as test sets. 

To assess paper reading performance, we employ following two measures: 
\begin{itemize}
    \item Factuality:  evaluates the accuracy of the system's response to factual information; 
    \item Informativeness: assesses the completeness of the system's response.
\end{itemize}


To evaluate paper polishing and academic translation performance, we use three criteria: 
\begin{itemize}
    \item Fluency: assesses the language coherence of model's outputs; 
    \item Fidelity: measures content faithfulness to the original text; 
    \item Academic: evaluates adherence to academic language standards.
\end{itemize}


\subsection{Results}

\input{reading_results}


The results of the paper reading are shown in Table 1. The highest results in the table are highlighted in bold, and the second-highest results are underlined.
SparkRA outperforms other models across all metrics. It achieves the highest score in Factuality with a score of 4.68, surpassing the closest competitor, GPT-4, which scores 4.67. In terms of Informativeness, SparkRA attains a score of 4.45, again leading over GPT-4, which scores 4.43. Overall, SparkRA achieves the highest average score of 4.57, demonstrating superior performance compared to other models like Llama3-8B. These results underscore SparkRA's effectiveness in producing factually accurate and informative text, establishing it as a state-of-the-art model in the paper reading task.

Table 2 shows the results of the paper polishing task. While Llama2-13B generates coherent text, it struggles with fidelity due to non-existent elements. Although Llama3-8B performs well across tasks, our SparkRA model, pre-trained on scientific literature and fine-tuned with 13 billion parameters, shows even greater improvement. SparkRA achieves state-of-the-art results compared to widely used LLMs like GPT-3.5 and GPT-4 across all evaluation metrics, excelling particularly in academic relevance.

\input{Polish_results}


Table 3 presents the academic translation results. SparkRA excels with the highest fidelity score (4.91) and the second-highest academic quality (4.75), showcasing its superior ability to preserve meaning and produce contextually appropriate translations. Additionally, SparkRA's BLEU score of 0.198 reflects its robustness in both human and automatic evaluations. Despite lower human evaluation scores than GPT-4, SparkRA's 13B parameter size offers flexibility, ease of training, and cost-effectiveness.

\input{Translate_results}

%% file: reading_results.tex
\begin{table}

    \begin{center}
    \resizebox{0.46\textwidth}{!}{
        \begin{tabular}{l|c|c|c}
            \hline \bf   \bf   & \bf Factuality  & \bf Informativeness & \bf Avg. \\
            \hline
Llama2-7B & 3.98 & 3.50 & 3.74 \\
Llama2-13B & 4.47 & 3.72 & 4.10 \\
Llama3-8B & 4.63 & 4.19 & 4.41 \\
\hline
GPT-3.5  & 4.20 & 3.97 & 4.09 \\
GPT-4 & \underline{4.67} & \underline{4.43} & \underline{4.55}\\
\hline
SparkRA & \textbf{4.68} & \textbf{4.45} & \textbf{4.57}\\
\hline
\end{tabular}}
\end{center}
\caption{\label{font-table} Results of paper reading task. 
}
\end{table}

%% file: Polish_results.tex
\begin{table}

    \begin{center}
        \resizebox{0.46\textwidth}{!}{
        \begin{tabular}{l|c|c|c|c}
            \hline   & \bf Fluency  & \bf Fidelity  & \bf Academic & \bf Avg.  \\
            \hline
Llama2-7B & \textbf{4.59} & 3.94 & 4.44 & 4.32 \\
Llama2-13B & \textbf{4.59} & 3.53 & 4.06 & 4.06 \\
Llama3-8B & \underline{4.56} & 3.97 & 4.47 & 4.33 \\
\hline
GPT-3.5  & 4.26 & 4.23 & 4.38 & 4.29\\
GPT-4 & 4.26 & \underline{4.29} & \underline{4.41} & \underline{4.32}\\
\hline
SparkRA & 4.41 & \textbf{4.45} & \textbf{4.61} & \textbf{4.49}\\
\hline
\end{tabular}}
\end{center}
\caption{\label{font-table} Results of paper polishing task. 
}
\end{table}

%% file: Translate_results.tex
\begin{table}

    \begin{center}
       \resizebox{0.46\textwidth}{!}{
        \begin{tabular}{l|c|c|c|c||c}
            \hline   & \bf Fluency  & \bf Fidelity  & \bf Academic & \bf Avg. & \bf BLEU \\
            \hline
Llama2-7B & 4.53 & 3.93 & 4.13 & 4.20 & 0.104 \\
Llama2-13B & \textbf{4.73} & 4.03 & 4.33 & 4.36 & 0.116 \\
Llama3-8B & \underline{4.64} & 4.46 & 4.43 & 4.51 & 0.168 \\
\hline
GPT-3.5  & 4.41 & 4.75 & 4.54 & 4.57 & 0.193 \\
GPT-4 & 4.50 & \underline{4.88} & \textbf{4.84} & \textbf{4.74} & 0.180 \\
\hline
SparkRA & 4.34 & \textbf{4.91} & \underline{4.75} & \underline{4.67} & \underline{0.198}\\
\hline
\end{tabular}}
\end{center}
\caption{\label{font-table} Results of academic translation task. 
}
\end{table}

%% file: related_work.tex
\section{Related Work}

\paragraph{Scientific literature pre-trained language model} Since the release of the pre-trained models \cite{vaswani2017transformer,radford2018gpt,devlin2019bert}, the language models for scientific literature have attracted the attention of scholars. 
These models are trained on various scientific datasets, with SciBERT on PubMed Central \cite{Beltagy2019scibert}, BioBERT and BioMegatron on biomedical literature \cite{Lee2020biobert,Shin2020biomegatron}, Galactica on multilingual articles \cite{Taylor2022galactica}, and ScholarBERT on ACL Anthology Corpus \cite{Hong2022scholarbert}.


\paragraph{Retrieval augmented generation to LLM} Retrieval-Augmented Generation (RAG), introduced by \citet{Lewis2020rag}, mitigates hallucinations in Large Language Models (LLMs) by integrating external data. \citet{Ma2023query} advanced RAG with query rewriting, while \citet{Chen2023benchmarking} benchmarked its effects, creating the RGB. \citet{Lyu2023improving} developed an algorithm for assessing retrieved data significance.


\paragraph{AI for science} Artificial intelligence has significantly impacted scientific research, enhancing efficiency and literature growth \cite{merchant2023nature,szymanski2023nature}.  
\citet{wang2023nature} proposed an AI-based scientific research method that can automatically extract useful information from a large amount of data and then use this information to conduct scientific research and discovery. 
Artificial intelligence technology has great potential in scientific research and discovery.

%% file: conclusion.tex
\section{Conclusion}


The SparkRA system, built on the SciLit-LLM, provides a comprehensive solution for academic tasks, including literature investigation, paper reading, and academic writing. Through extensive experiments, SparkRA demonstrated superior performance compared to existing models like ChatGPT, and even surpassed GPT-4 in specific tasks such as paper polishing, demonstrating its potential to enhance productivity for researchers and students with its precise and context-aware support for academic activities.

